%% file: paper.tex
\documentclass[usletterpaper, 10 pt, conference]{ieeeconf}

\IEEEoverridecommandlockouts %

\usepackage{amsmath,amsfonts,amssymb}
\usepackage{amsfonts} %
\usepackage{mathrsfs} %
\usepackage{bbold} %
\usepackage{pgfplots}
\usepackage{subfig}
\usepackage{tabularx}
\usepackage{multirow}
\usepackage{siunitx}
\usepackage{listings}
\usepackage{flushend}
\usepackage{algorithm}
\usepackage{algpseudocode}
\usepackage[export]{adjustbox}

\usepackage{caption}
\usepackage{graphicx}
\usepackage{transparent}
\graphicspath{{figures/}}

\usepackage{xargs} %

\usepackage[style=ieee, backend=biber, bibencoding=utf8, maxbibnames=5, mincitenames=1, maxcitenames=2, url=false, doi=false, isbn=false, citestyle=numeric-comp, natbib=true, eprint=false]{biblatex}

\usepackage[colorinlistoftodos,prependcaption,textsize=small]{todonotes}
\graphicspath{ {figures/} }

\newcommandx{\change}[2][1=]{\todo[inline,linecolor=blue,backgroundcolor=blue!25,bordercolor=blue,#1]{#2}}

\usepackage{bm} %
\usepackage{lipsum}
 
\newsavebox{\measuredSize}

\newcommand\copyrighttext{%
\scriptsize \textcolor{blue}{\textcopyright 2020 IEEE. Accepted at IROS 2020 - Workshop PLC. Personal use of this material is permitted. Permission from IEEE must be obtained for all other uses, in any current or future media, including reprinting/republishing this material for advertising or promotional purposes, creating new collective works, for resale or redistribution to servers or lists, or reuse of any copyrighted component of this work in other works}} \newcommand\copyrightnotice{%
\begin{tikzpicture}[remember picture,overlay] \node[anchor=north,yshift=-7.5pt] at (current page.north) {\fbox{\parbox{\dimexpr\textwidth-\fboxsep-\fboxrule\relax}{\copyrighttext}}}; \end{tikzpicture}%
}

\pgfplotsset{compat=newest}
\pgfplotsset{plot coordinates/math parser=false}
\usetikzlibrary{decorations.markings, positioning}

\newlength\figureheight 
\newlength\figurewidth 

\title{\LARGE \bf Counterfactual Policy Evaluation for Decision-Making in Autonomous Driving}
\author{Patrick Hart$^{1}$ and Alois Knoll$^{2}$%
	\thanks{$^{1}$fortiss GmbH, An-Institut Technische Universit\"{a}t M\"{u}nchen, Munich, Germany}%
	\thanks{$^{2}$Alois Knoll is with Robotics, Artificial Intelligence and Real-time Systems, Technische Universit\"{a}t M\"{u}nchen, Munich, Germany}%
}

\DeclareTextFontCommand{\tsf}{\tiny\sffamily} %

\usepackage{listings}
\usepackage{xcolor}

\colorlet{punct}{red!60!black}
\definecolor{background}{HTML}{EEEEEE}
\definecolor{delim}{RGB}{50,105,176}
\colorlet{numb}{magenta!60!black}

\lstdefinelanguage{json}{
    basicstyle=\normalfont\ttfamily,
    numbers=none,
    stepnumber=1,
    showstringspaces=false,
    breaklines=true,
    frame=lines,
    backgroundcolor=\color{background},
    literate=
     *{0}{{{\color{numb}0}}}{1}
      {1}{{{\color{numb}1}}}{1}
      {2}{{{\color{numb}2}}}{1}
      {3}{{{\color{numb}3}}}{1}
      {4}{{{\color{numb}4}}}{1}
      {5}{{{\color{numb}5}}}{1}
      {6}{{{\color{numb}6}}}{1}
      {7}{{{\color{numb}7}}}{1}
      {8}{{{\color{numb}8}}}{1}
      {9}{{{\color{numb}9}}}{1}
      {:}{{{\color{punct}{:}}}}{1}
      {,}{{{\color{punct}{,}}}}{1}
      {\{}{{{\color{delim}{\{}}}}{1}
      {\}}{{{\color{delim}{\}}}}}{1}
      {[}{{{\color{delim}{[}}}}{1}
      {]}{{{\color{delim}{]}}}}{1},
}

\begin{document}

\maketitle
\copyrightnotice
\thispagestyle{empty}
\pagestyle{empty}

\global\csname @topnum\endcsname 0
\global\csname @botnum\endcsname 0

\newcommand{\figurename}{Fig. }

\begin{abstract}
\input{abstract}
\end{abstract}

\IEEEpeerreviewmaketitle

\input{introduction}
\input{related_work}

\input{approach}

\input{evaluation}
\input{conclusion}

\section*{Acknowledgment}
This research was funded by the Bavarian Ministry of Economic Affairs, Regional Development and Energy, and by the project Dependable AI.

\section*{Appendix}
\label{sec:appendix}
Parameters used for learning the ego vehicle's behavior policy and other parameters used in the simulation:
\begin{lstlisting}[language=json,firstnumber=1]
"BehaviorIDM": {
  "MaxVelocity": 30.0,
  "MinimumSpacing": 2.0,
  "DesiredTimeHeadway": [1.0, 5.0],
  "MaxAcceleration": 2.5,
  "DesiredVelocity": 15.0,
  "ComfortableBrakingAcceleration": 1.6,
  "MinVelocity": 0.0,
  "Exponent": 4
},
"BehaviorSACAgent": {
  "ActorFcLayerParams": [512,256,256],
  "CriticJointFcLayerParams": [512,256,256],
  "ActorLearningRate": 0.0003,
  "CriticLearningRate": 0.0003,
  "AlphaLearningRate": 0.0003,
  "TargetUpdateTau": 0.05,
  "TargetUpdatePeriod": 3,
  "Gamma": 0.995,
  "RewardScaleFactor": 1.0,
  "ReplayBufferCapacity": 10000,
  "BatchSize": 512,
},
"step_time": 0.2
\end{lstlisting}

\renewcommand{\bibfont}{\small}
\printbibliography

\end{document}

%% file: abstract.tex
Learning-based approaches, such as reinforcement and imitation learning are gaining popularity in decision-making for autonomous driving.
However, learned policies often fail to generalize and cannot handle novel situations well.
Asking and answering questions in the form of “Would a policy perform well if the other agents had behaved differently?” can shed light on whether a policy has seen similar situations during training and generalizes well.
In this work, a counterfactual policy evaluation is introduced that makes use of counterfactual worlds --- worlds in which the behaviors of others are non-actual.
If a policy can handle all counterfactual worlds well, it either has seen similar situations during training or it generalizes well and is deemed to be fit enough to be executed in the actual world.
Additionally, by performing the counterfactual policy evaluation, causal relations and the influence of changing vehicle's behaviors on the surrounding vehicles becomes evident.
To validate the proposed method, we learn a policy using reinforcement learning for a lane merging scenario.
In the application-phase, the policy is only executed after the counterfactual policy evaluation has been performed and if the policy is found to be safe enough.
We show that the proposed approach significantly decreases the collision-rate whilst maintaining a high success-rate.

%% file: introduction.tex
\section{Introduction}
\label{sec:Introduction}
Learning-based approaches become ever more prevalent in the decision-making community for autonomous driving.
Often, deep neural networks (DNNs) are used in various approaches to learn behaviors or to predict other traffic participants.
Popular approaches for learning behaviors include reinforcement learning \cite{shalevshwartz2016safe, hart2020graph}, imitation learning \cite{ho2016generative, li2017infogail, bansal2018chauffeurnet}, and other methodologies \cite{bojarski2016end, codevilla2018end}.
These approaches often share one common characteristic --- they use DNNs as function approximators for their policy.
However, DNNs often tend to overfit \cite{zhang2018study, song2019observational} or do not generalize well to novel situations \cite{kawaguchi2017generalization, cobbe2018quantifying}.
Executing learned policies in safety-critical applications, such as autonomous driving without any estimate of their performance poses significant threats.
For example, would the learned policies still perform well if the leading vehicle would suddenly accelerate or decelerate?

\begin{figure}[!t]
	\vspace{0.15cm}
	\footnotesize
	\centering
	\def\svgwidth{\columnwidth}
	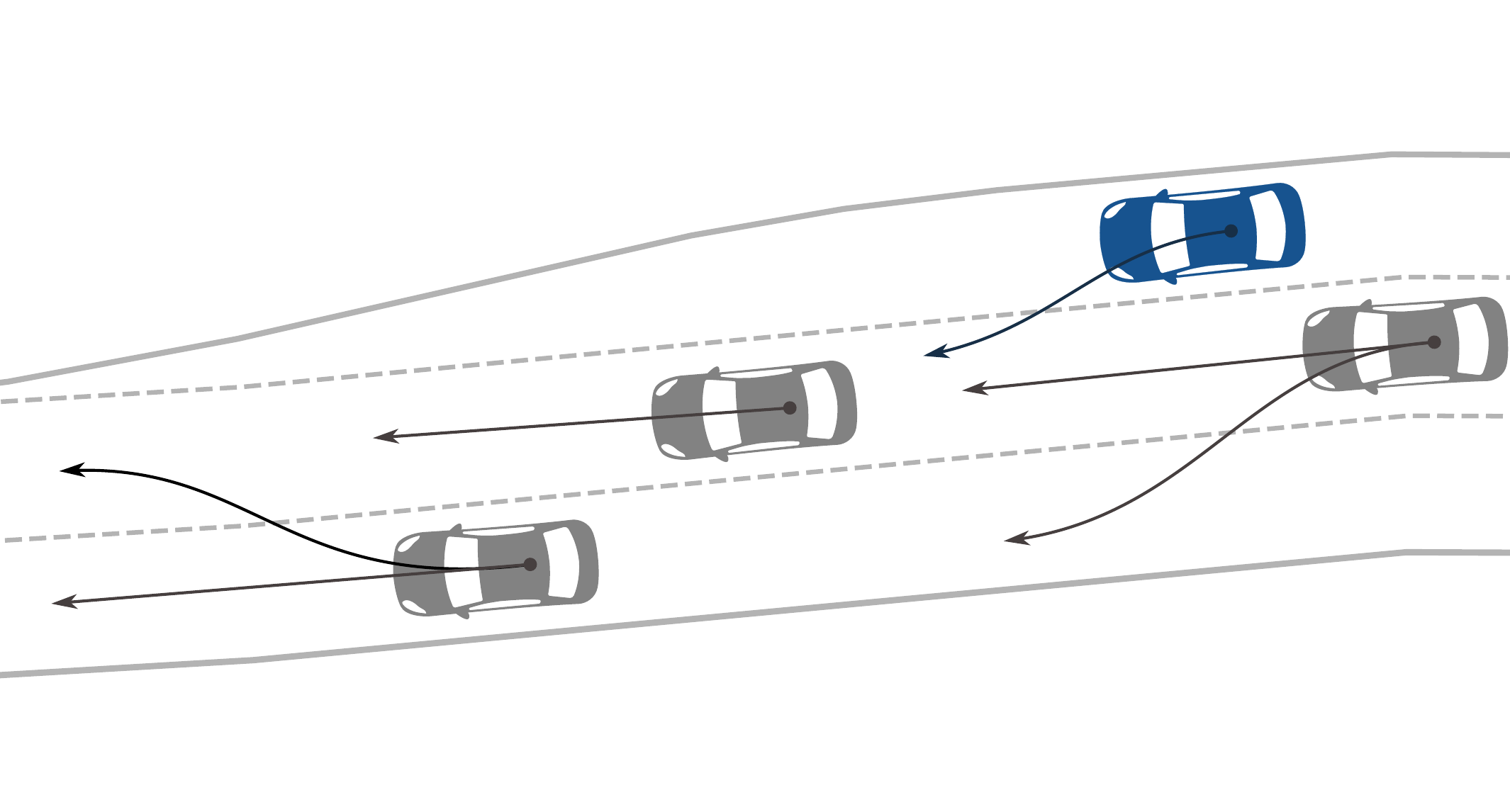
  \caption{The behaviors $\pi_{1:N}$ of other traffic participants are not observable by the ego vehicle (depicted in blue) resulting in an uncertain environment.
		   Vehicles can suddenly change their driving behavior, such as braking or changing the lane.
		   Would the policy still perform well and be safe if the other vehicle changed their behavior?
		}
	\label{fig:intro}
	\vspace{-0.15cm}
\end{figure}

In this work, a counterfactual policy evaluation (CPE) is introduced that evaluates the performance of a learned behavior policy $\pi$ prior to its execution in the actual world using counterfactual worlds.
In a counterfactual world, the behavior policy of at least one other vehicle is replaced and, thus, non-actual behaviors are introduced.
Using these worlds, counterfactual reasoning can be performed to answer questions, such as: Would a policy perform well if the other vehicles had behaved as in the counterfactual world?
Performing counterfactual reasoning is especially important in applications, such as autonomous driving, where the behavior of others is not directly observable and, thus, uncertain.
To perform CPE, counterfactual worlds are generated by replacing the behavior policies of vehicles surrounding the ego vehicle.
To determine and better estimate the direct effect of replacing behavior policies on the policy under test (PUT), the replaced behaviors are chosen to be independent of their surroundings --- they do not interact with or react to their surroundings.
By simulating the counterfactual worlds into the future and by evaluating the evolution of these, we can obtain estimates on how the PUT would perform if the other vehicles behave as in the counterfactual worlds.
If the PUT has seen similar situations as in the counterfactual worlds during training or if it generalizes well, it should be able to handle all counterfactual worlds sufficiently well.
Thus, in this work, the PUT is only deemed safe enough to be executed if there are no or only a few collisions in the counterfactual worlds.
We define an upper collision-rate threshold that defines whether the PUT shall be executed in the actual world posterior to its CPE.
Additionally, by using CPE, we can extract causal relations and make the influence of changing other vehicle's behavior policies become evident.
Especially when using black-box approaches, such as DNNs, causal relations and how the PUT reacts to changes in its surroundings are important to observe.
For example, if the PUT just had memorized certain situations during training it would not react to changes in its environment and most probably would fail to generalize to and handle novel situations.

To validate the proposed approach and to empirically show the merits of performing CPE, we use a lane merging scenario.
First, we learn a policy $\pi_{ego}$ using an actor-critic reinforcement learning algorithm.
In the application-phase, the learned policy $\pi_{ego}$ becomes the PUT and is only being executed after the CPE has been performed and is deemed safe enough to be executed.
Otherwise, the ego vehicle is controlled by a conventional lane-following behavior policy.
We show that performing CPE prior to the execution in the actual world of the learned policy significantly decreases the collision-rate whilst maintaining high performance in terms of reaching the vehicle's goal.

This work is further structured as follows: Section \ref{sec:RelatedWork} gives an overview of related work in counterfactual reasoning.
Section \ref{sec:Approach} outlines the proposed approach in detail.
In Section \ref{sec:Experiments}, the results and findings are presented, and finally, a conclusion is given.

%% file: figures/intro.pdf_tex
\begingroup%
  \makeatletter%
  \providecommand\color[2][]{%
    \errmessage{(Inkscape) Color is used for the text in Inkscape, but the package 'color.sty' is not loaded}%
    \renewcommand\color[2][]{}%
  }%
  \providecommand\transparent[1]{%
    \errmessage{(Inkscape) Transparency is used (non-zero) for the text in Inkscape, but the package 'transparent.sty' is not loaded}%
    \renewcommand\transparent[1]{}%
  }%
  \providecommand\rotatebox[2]{#2}%
  \newcommand*\fsize{\dimexpr\f@size pt\relax}%
  \newcommand*\lineheight[1]{\fontsize{\fsize}{#1\fsize}\selectfont}%
  \ifx\svgwidth\undefined%
    \setlength{\unitlength}{613.37141382bp}%
    \ifx\svgscale\undefined%
      \relax%
    \else%
      \setlength{\unitlength}{\unitlength * \real{\svgscale}}%
    \fi%
  \else%
    \setlength{\unitlength}{\svgwidth}%
  \fi%
  \global\let\svgwidth\undefined%
  \global\let\svgscale\undefined%
  \makeatother%
  \begin{picture}(1,0.53791691)%
    \lineheight{1}%
    \setlength\tabcolsep{0pt}%
    \put(0,0){\includegraphics[width=\unitlength,page=1]{intro.pdf}}%
    \put(0.65814584,0.46888997){\color[rgb]{0,0,0}\makebox(0,0)[lt]{\lineheight{1.25}\smash{\begin{tabular}[t]{l}$\pi_{ego}$\end{tabular}}}}%
    \put(0,0){\includegraphics[width=\unitlength,page=2]{intro.pdf}}%
    \put(0.3470364,0.02026449){\color[rgb]{0,0,0}\makebox(0,0)[lt]{\lineheight{1.25}\smash{\begin{tabular}[t]{l}$\pi_2$\end{tabular}}}}%
    \put(0.38337592,0.41031765){\color[rgb]{0,0,0}\makebox(0,0)[lt]{\lineheight{1.25}\smash{\begin{tabular}[t]{l}$\pi_1$\end{tabular}}}}%
    \put(0.88335325,0.17933067){\color[rgb]{0,0,0}\makebox(0,0)[lt]{\lineheight{1.25}\smash{\begin{tabular}[t]{l}$\pi_0$\end{tabular}}}}%
    \put(0,0){\includegraphics[width=\unitlength,page=3]{intro.pdf}}%
  \end{picture}%
\endgroup%

%% file: related_work.tex
\section{Related Work}
\label{sec:RelatedWork}
In this section, we provide a brief overview of counterfactual reasoning and counterfactual decision-making.

The term counterfactual reasoning has predominantly been coined by Judea Pearl \cite{pearl2018book} who also introduced a mathematical framework (do-calculus) for counterfactual reasoning \cite{pearl2012docalculus}.
Do-calculus aims to describe the human capacity to reason about counterfactual outcomes of past experiences with the goal of “mining worlds that could have been”.
In safety-critical applications, such as autonomous driving do-calculus often is the only option to obtain certain distributions.
For example, the probability distribution $P(\text{collision} | \pi_{other}=\pi_{0})$ cannot directly be collected in the actual world.
Having a joint probability distribution, causal assumptions, and using do-calculus this distribution can be inferred a-posteriori.
In this work, contrary to using a joint probability distribution and causal assumptions, we can directly collect these probability distributions in simulation and do not have to infer these.

Another approach in the literature is counterfactual explanations where causal situations in the form of "If X had not occurred, Y would not have occurred" are formulated and answered \cite{molnar2020interpretable}.
In the context of autonomous driving, this is can be re-formulated as e.g.\ "If the other vehicle had not accelerated, a collision with the ego-vehicle would not have occurred".
Using the in this work defined counterfactual worlds, we can infer which event lead to a certain outcome -- e.g. if a collision could be avoided if the other vehicle had decelerated or accelerated.

In literature, there exist various approaches that make use of counterfactuals for decision-making.
\citet{Buesing2019} propose a counterfactual guided-policy search reinforcement learning algorithm.
In their work, they leverage a model to consider alternative outcomes and, thus, increase the algorithms sample efficiency on required experiences.
\citet{Isele2019} consider counterfactual behaviors in the prediction by spanning an intention tree.
Counterfactual regret minimization (CRM) has been widely used for finding best response strategies in multi-player games \cite{brown2019deep, neller2013introduction}.
In CRM, the choice of future actions is based on the regret the agent accumulates of not having chosen specific actions and self-play.

%% file: approach.tex
\section{Approach}
\label{sec:Approach}
\begin{figure}[!t]
	\vspace{0.15cm}
	\footnotesize
	\centering
	\def\svgwidth{\columnwidth}
	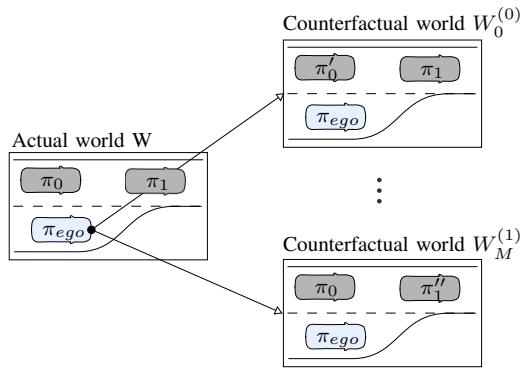
  \caption{Counterfactual worlds $W^{(\cdot)}_i$ are created by mirroring the actual world $W$ and by exchanging the vehicle's policies.
           In the counterfactual world $W^{(0)}_0$ the policy of vehicle $v_0$ has been exchanged and in counterfactual world $W^{(1)}_M$ the policy of vehicle $v_1$.}
	\label{fig:intro}
	\vspace{-0.15cm}
\end{figure}

This section describes the counterfactual policy evaluation (CPE) algorithm in detail.
Using CPE, the performance of executing a policy prior to execution in the actual world is being evaluated using counterfactual reasoning.
First, we introduce all necessary tools and notation as a basis and then outline the full structure of the algorithm.

In each world $W$, there is a set of vehicles $V=(v_{0}, v_{j}, \dots, v_{N})$ with $N$ being the number of vehicles.
Each vehicle $v_j$ has a policy $\pi_j$ that specifies its behavior.
The full parameterization of a world $W$ is given by a set of vehicles paired with their respective policies $\pi_j$.
A policy $\pi_j$ outputs an action $a$ for each world state $W$.
A full world state is given by $W=((v_{0}, \pi_0), (v_{ego}, \pi_{ego}), \dots, (v_N, \pi_N))$ with one of the $N$ vehicles being the ego vehicle $v_{ego}$ and that is being controlled by the ego policy $\pi_{ego}$.
The ego vehicle's policy $\pi_{ego}$ is defined to be the policy under test (PUT).
We neglect the map and environment information in the notation as these are assumed to be the same in all worlds.

In a counterfactual world, the policy of at least one vehicle is replaced by an independent policy.
To determine the direct effect of replacing an individual policy on its surrounding, the replaced behavior policies are independent -- they do not react to or interact with their surroundings.
For example, policies that accelerate or decelerate independent of changes in their surroundings.
The behavior policies of the $K$ nearest vehicles are being replaced as we assume that vehicles far away do not have a large direct effect on the PUT.
These independent behavior policies are assigned from a pre-defined policy pool $\pi_{0}, \dots, \pi_{M} \sim \mathcal{P}$ to vehicles other than the ego vehicle resulting in $K \times M$ counterfactual worlds.
A counterfactual world where the policy of vehicle $v_j$ has been replaced by the independent policy $\pi_i$ is denoted as $W^{(j)}_i$.
We introduce a short form notation $W^{(j)}_{\mathcal{P}}$ for the set of counterfactual worlds that are generated by replacing the behavior policy of vehicle $v_j$ $M$ times. 
These worlds are called counterfactual worlds, as we assume non-actual behavior policies for other vehicles.
This allows us to answer counterfactual questions, such as: How would the PUT perform if the other vehicles had behaved as in the counterfactual world?

To perform counterfactual reasoning, the set of counterfactual worlds $W^{(j)}_\mathcal{P}$ is forward simulated for a pre-defined time-horizon $T_{ct}$ starting at the current world time $t_w$.
From the forward simulation trajectories $\tau^{(j)}_\mathcal{P}$ of tuples in the form of $\langle observation, terminal, reward, info \rangle$ are collected within the time-range $[t_w, t_w + T_{ct}]$ where the polilcy of vehicle's $v_j$ has been replaced by the independent behavior policy $\pi_i$.

Using the recorded future trajectories $\tau^{(j)}_\mathcal{P}$ of the counterfactual worlds, conditional probabilities $P(X | W^{(j)}_\mathcal{P})$ for the PUT can be obtained.
The conditional probability $P(X | {W}^{(j)}_\mathcal{P})$ for exchanging vehicle's $v_j$ policy is given by
\begin{equation}
  P(X | {W}^{(j)}_\mathcal{P}) = \frac{1}{M} \sum_{i=0}^{M} \text{I}\_(\tau^{(j)}_i)
  \label{eq:prob_dist}
\end{equation}
with a boolean indicator function $\text{I}\_(x)$ returning whether a pre-defined event, such as a colission has occured in the trajectory $\tau^{(j)}_i$.

In safety-critical applications, such as autonomous driving, we especially are interested in the average conditional collisions probability over all counterfactual worlds denoted as $P(C | W^{(\cdot)}_\mathcal{P})$.
For convenience of notation, we define the average conditional collision probability as 
\begin{equation}
  P_\text{C} := \frac{1}{K} \sum_{j=0}^{K} P(C | {W}^{(j)}_\mathcal{P})
  \label{eq:col_rate}
\end{equation}
with $N$ being the number of vehicles.

We use the average conditional collision probability $P_\text{C}$ of the counterfactual worlds as decision criterion on whether the PUT should be executed in the actual world $W$.
We define an upper collision threshold $\rho_\text{max}$ up to which the PUT is being executed.
If $P_\text{C}$ is larger, the ego vehicle is controlled by a lane-following model, such as the intelligent driver model (IDM) instead \cite{Treiber2000}.
We assume that the lane-following model is collision-free when driving on a single lane.
The full CPE algorithm is outlined in Algorithm \ref{alg:cpe}.

Additionally, due to the replaced behavior policies being indepent of their surroundings, causal relations between the vehicle's policies can be extracted.
In this work, we use the mean average displacement (MAD) to measure the distance of the vehicle's states in the counterfactual worlds $W^{(j)}_\mathcal{P}$ to the actual world $W$.
The greater the MAD in a counterfactual world is compared to the actual world, the greater the influence of changing the other vehicle's policy is on the PUT.
This indicates how well and if the PUT reacts to its surrounding and changes that occur in it.
For example, if the leading vehicle of the ego vehicle brakes rapidly and the PUT does not react at all, this could be an indicator that the policy does not generalize well and that it did not see similar behaviors during training.

\begin{algorithm}[t]
  \caption{Counterfactual Policy Evaluation (CPE)}
  \label{alg:cpe}
  \begin{algorithmic}
  \Function{CPE}{World $W$, PolicyPool $\mathcal{P}$}
  \State $v_{0:K}$ = NearbyVehicles(W)
  \State trajectories = []
  \For{v in $v_{0:K}$}
    \State $W^{(v)}_{\mathcal{P}}$ = GetCounterfactualWorlds(W, $v$, $\mathcal{P}$)
    \State $\tau^{(v)}_{\mathcal{P}}$ = ForwardSimulate($W^{(v)}_{\mathcal{P}}$, $T_{ct}$)
    \State trajectories.append($\tau^{(v)}_{\mathcal{P}}$)
  \EndFor
  \State return ToBeExecuted(trajectories, $\rho_{max}$)
  \EndFunction
  \end{algorithmic}
\end{algorithm}

In the next section, we present results and an evaluation showing that the CPE algorithm significantly descreases the colllision-rate and increases the performance of executing a learned policy in the actual world.
Additionally, we conduct studies on varying the maximum allowed collision-rate $\rho_{max}$.

%% file: figures/virtual_worlds.pdf_tex
\begingroup%
  \makeatletter%
  \providecommand\color[2][]{%
    \errmessage{(Inkscape) Color is used for the text in Inkscape, but the package 'color.sty' is not loaded}%
    \renewcommand\color[2][]{}%
  }%
  \providecommand\transparent[1]{%
    \errmessage{(Inkscape) Transparency is used (non-zero) for the text in Inkscape, but the package 'transparent.sty' is not loaded}%
    \renewcommand\transparent[1]{}%
  }%
  \providecommand\rotatebox[2]{#2}%
  \newcommand*\fsize{\dimexpr\f@size pt\relax}%
  \newcommand*\lineheight[1]{\fontsize{\fsize}{#1\fsize}\selectfont}%
  \ifx\svgwidth\undefined%
    \setlength{\unitlength}{523.27932763bp}%
    \ifx\svgscale\undefined%
      \relax%
    \else%
      \setlength{\unitlength}{\unitlength * \real{\svgscale}}%
    \fi%
  \else%
    \setlength{\unitlength}{\svgwidth}%
  \fi%
  \global\let\svgwidth\undefined%
  \global\let\svgscale\undefined%
  \makeatother%
  \begin{picture}(1,0.61619588)%
    \lineheight{1}%
    \setlength\tabcolsep{0pt}%
    \put(-0.21894307,0.70802863){\color[rgb]{0,0,0}\makebox(0,0)[lt]{\begin{minipage}{1.43122132\unitlength}\raggedright \end{minipage}}}%
    \put(0.54037709,0.56185979){\color[rgb]{0,0,0}\makebox(0,0)[lt]{\lineheight{1.25}\smash{\begin{tabular}[t]{l}Counterfactual world $W^{(0)}_0$\end{tabular}}}}%
    \put(0.54037709,0.22142967){\color[rgb]{0,0,0}\makebox(0,0)[lt]{\lineheight{1.25}\smash{\begin{tabular}[t]{l}Counterfactual world $W^{(1)}_M$\end{tabular}}}}%
    \put(0.12276278,0.381819){\color[rgb]{0,0,0}\makebox(0,0)[lt]{\lineheight{1.25}\smash{\begin{tabular}[t]{l}Actual world W\end{tabular}}}}%
    \put(0,0){\includegraphics[width=\unitlength,page=1]{virtual_worlds.pdf}}%
    \put(0.16708055,0.32132986){\color[rgb]{0,0,0}\makebox(0,0)[lt]{\lineheight{1.25}\smash{\begin{tabular}[t]{l}$\pi_0$\end{tabular}}}}%
    \put(0.1698493,0.24692117){\color[rgb]{0,0,0}\makebox(0,0)[lt]{\lineheight{1.25}\smash{\begin{tabular}[t]{l}$\pi_{ego}$\end{tabular}}}}%
    \put(0,0){\includegraphics[width=\unitlength,page=2]{virtual_worlds.pdf}}%
    \put(0.58916444,0.49473376){\color[rgb]{0,0,0}\makebox(0,0)[lt]{\lineheight{1.25}\smash{\begin{tabular}[t]{l}$\pi_0'$\end{tabular}}}}%
    \put(0.75184875,0.49371698){\color[rgb]{0,0,0}\makebox(0,0)[lt]{\lineheight{1.25}\smash{\begin{tabular}[t]{l}$\pi_1$\end{tabular}}}}%
    \put(0.59193324,0.42032507){\color[rgb]{0,0,0}\makebox(0,0)[lt]{\lineheight{1.25}\smash{\begin{tabular}[t]{l}$\pi_{ego}$\end{tabular}}}}%
    \put(0,0){\includegraphics[width=\unitlength,page=3]{virtual_worlds.pdf}}%
    \put(0.58916444,0.15668583){\color[rgb]{0,0,0}\makebox(0,0)[lt]{\lineheight{1.25}\smash{\begin{tabular}[t]{l}$\pi_0$\end{tabular}}}}%
    \put(0.75184875,0.15566906){\color[rgb]{0,0,0}\makebox(0,0)[lt]{\lineheight{1.25}\smash{\begin{tabular}[t]{l}$\pi_1''$\end{tabular}}}}%
    \put(0.59193324,0.08227716){\color[rgb]{0,0,0}\makebox(0,0)[lt]{\lineheight{1.25}\smash{\begin{tabular}[t]{l}$\pi_{ego}$\end{tabular}}}}%
    \put(0,0){\includegraphics[width=\unitlength,page=4]{virtual_worlds.pdf}}%
    \put(0.32976527,0.32031309){\color[rgb]{0,0,0}\makebox(0,0)[lt]{\lineheight{1.25}\smash{\begin{tabular}[t]{l}$\pi_1$\end{tabular}}}}%
    \put(0,0){\includegraphics[width=\unitlength,page=5]{virtual_worlds.pdf}}%
  \end{picture}%
\endgroup%

%% file: evaluation.tex
\section{Experiments}
\label{sec:Experiments}
In this section, experiments are performed and an evaluation of the counterfactual policy evaluation (CPE) is presented using a highway merging senario.
First, we introduce the used simulation framework and the scenario.
Next, we outline the actor-critic reinforcement learning approach used for learning lane merging policies.
And finally, we apply the CPE to the learned policy and present results and findings.

\subsection{Simulation and Scenario}
\begin{figure}[!t]
	\vspace{0.15cm}
	\footnotesize
	\centering
	\def\svgwidth{\columnwidth}
	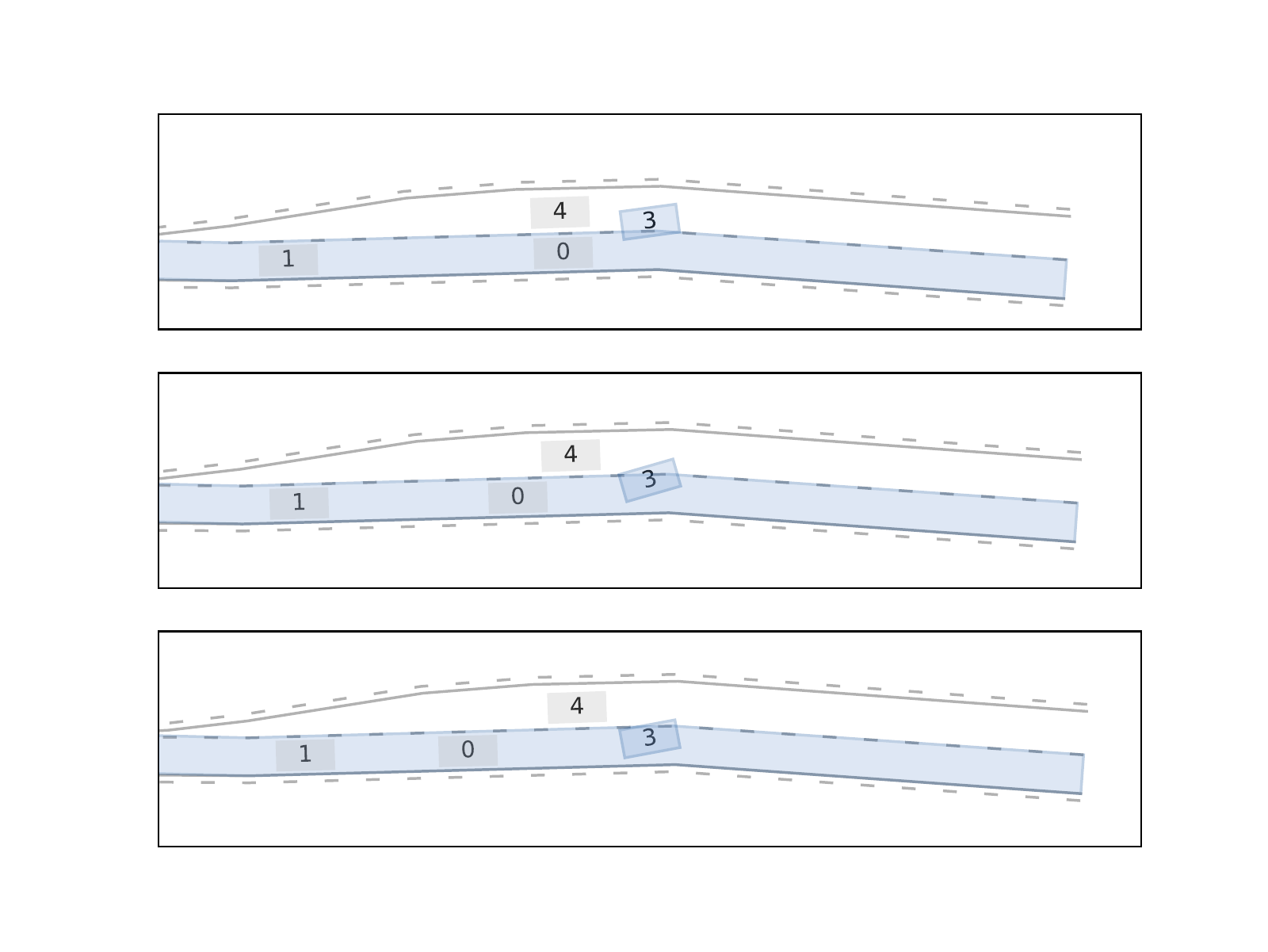
	\caption{Three counterfactual worlds $W^{(0)}_i$ are shown where the policy of vehicle $v_0$ has been replaced.
			 Initially, all these worlds had the same state but evolved differently over time.
			 On the top, vehicle $v_0$ decelerates, in the middle, vehicle $v_0$ drives with a constant velocity, and, on the bottom, vehicle $v_0$ accelerates.
			 The ego vehicle $v_3$ (depicted in blue) is controlled by the learned policy $\pi_\psi(a | s)$.
			 This figure shows the counterfactual world at time $t_w + 1s$.}
	\label{fig:virtual_worlds}
	\vspace{-0.15cm}
\end{figure}

We use the semantic simulation framework BARK\footnote{\url{https://github.com/bark-simulator/bark/}} and its machine learning extension BARK-ML\footnote{\url{https://github.com/bark-simulator/bark-ml/}} for all simulations and to learn the ego vehicle's policy $\pi_{ego}$.
To demonstrate and evaluate the approach, we chose a lane merging scenario where the ego vehicle's goal is to merge onto the highway.
The scenarios are generated using a sampling strategy that samples the initial conditions and the behavioral parameters of the other vehicles according to a pre-defined distribution.
All other vehicles are controlled by the ruled-based Mobil moddel.
The other vehicles are controlled by a rule-based model --- the Mobil model \cite{treiber2016mobil}.
Further simulation parameters are outlined in the Appendix.
The simulation is episodic and ends either if the goal has been reached, a collision has occurred, or the maximum number of steps has been reached.
The goal of the ego vehicle is defined on the left lane using a polygonal area (depicted in blue in Figure \ref{fig:virtual_worlds}) and the vehicle has to reach this area within a velocity range of $[5m/s, 16m/s]$ and the vehicle angle $\theta$ in the range of $[-0.05rad, 0.05rad]$.

\subsection{Learning Policies}
We base learning our policy on \cite{Hart2020} and use the soft actor-critic (SAC) \cite{Haarnoja2018} algorithm to learn a stochastic policy $\pi_\chi(a | s)$ that is parameterized by the neural network parameters $\chi$ for changing lanes.
The action space is comprised of the acceleration $a$ and the steering-rate $\delta$.
To generate an observation for a single time-step, we use a nearest state observer that concatenates the states of $l=5$ nearest vehicles (cartesian coordinates, velocity, and vehicle angle).
The ego vehicle's state is always in the first position in the observation and the other vehicles follow sorted based on their distance to the ego vehicle.
If there are less than five objects the rest of the observation $s$ is filled with zeros.
The reward signal $r_t$ for time $t$ is composed of several terms.
Reaching the goal is rewarded with a positive reward of $10$ and a collision is penalized with a reward of $-10$.
To accelerate the learning process, we further introduce a guiding reward signal that uses the $L2$ norm to the defined goal state (distance, deviation of the vehicle angle, and the deviation to the goal velocity).
To achieve more smooth and comfortable driving behaviors, we additionally put a sqared penality term on the actions.
All rewards besides the goal and collision term are a factor of $10$ smaller to prioritize to have no collisions and reaching the goal state.
In this work, we output a normal distribution $\underline{a} \sim \mathcal{N}(\underline{\mu}, \underline{\sigma})$ for the policy $\pi$ with $\underline{\mu}$ being the means and $\underline{\sigma}$ the standard deviations for the actions.
During training, this stochastic policy is being sampled to explore the configuration space and during application, the mean of each action is used.
The hyperparameters used for the SAC algorithm are outlined in the Appendix.

After $250,000$ thousand training episodes, a success-rate of $32\%$ and a collision-rate of $48\%$ is achieved.
Even after fine-tuning the hyperparameters and a longer training time, collisions persist.
Many machine learning approaches cannot guarantee or achieve a collision-rate of zero.
Therefore, they are considered to be infeasible for safety-critical situations, such as autonomous driving.
To decrease the collision-rate and to obtain better estimates on the performance of a policy $\pi$, this work introduces a counterfactual policy evaluation.

\begin{figure}[!t]
	\vspace{0.15cm}
	\footnotesize
	\centering
	\def\svgwidth{\columnwidth}
	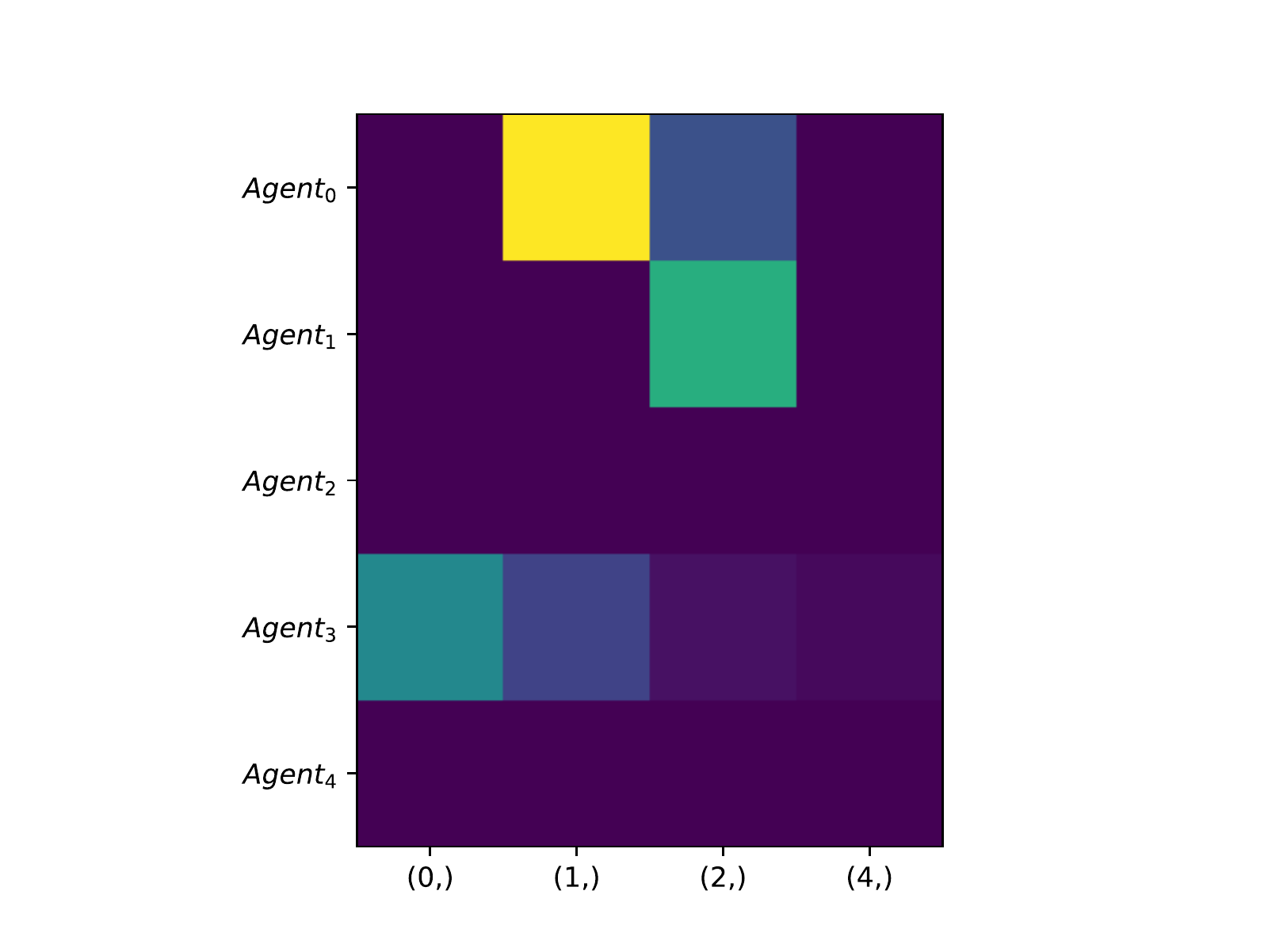
	\caption{The heatmap shows the mean average displacement of the vehicle's states over the counterfactual worlds $W^{(j)}_\mathcal{P}$.
			 Vehicle $v_3$ is controlled by the learned policy and the heatmap shows it being influenced by all surrounding vehicles.
			 The other vehicles are controlled by a lane-following model and, thus, are only influenced by their leading vehicles on the same lane.}
	\label{fig:influence_heatmap}
	\vspace{-0.15cm}
\end{figure}

\subsection{Counterfactual Policy Evaluation (CPE)}
For the CPE, we use a behavior policy pool $\mathcal{P}$ with a constant acceleration policy $\mathcal{B}$.
In this work, we use three different parameterizations for the constant acceleration policy $\mathcal{B}$ having the accelerations of $a = [-2, 0, 2]$ and resulting in three behavior policies.
Thus, resulting in three counterfactual worlds $(W^{(j)}_0, W^{(j)}_1, W^{(j)}_2)$ for each vehicle $v_j$, where the replaced vehicle decelerates, drives with constant velocity, and accelerates.
Further, we simulate each counterfactual world for a pre-defined horizon $T_{ct} = 1s$ from the current world time $t_w$.

Figure \ref{fig:virtual_worlds} shows the evolution of the counterfactual worlds $W^{(0)}_i$ at the time $t=t_w + 1s$ where the policy of vehicle $v_0$ has been replaced.
Initially, all worlds had the same initial state but then evolved differently due to the policy of vehicle $v_0$ being replaced by the policies from the behavior policy pool $\mathcal{P}$.
Vehicle $v_3$ is controlled by the learned SAC policy $\pi_\chi(a | s)$ and is also the policy under test (PUT).
In Figure \ref{fig:virtual_worlds} on the top, in counterfactual world $W^{(0)}_0$ vehicle $v_0$ decelerates, in the middle, vehicle $v_0$ drives with a constant velocity, and on the bottom vehicle $v_0$ accelerates.
In each counterfactual world, the evolution over time and the reaction of the PUT to the environment differs significantly.

As can be seen, the learned policy $\pi_\chi(a | s)$ is able to handle all of the counterfactual worlds well and does not cause any collisions.
Thus, resulting in the conditional collision probability of $P_C = 0$.

\begin{figure}[!t]
	\vspace{0.15cm}
	\footnotesize
	\centering
	\def\svgwidth{\columnwidth}
	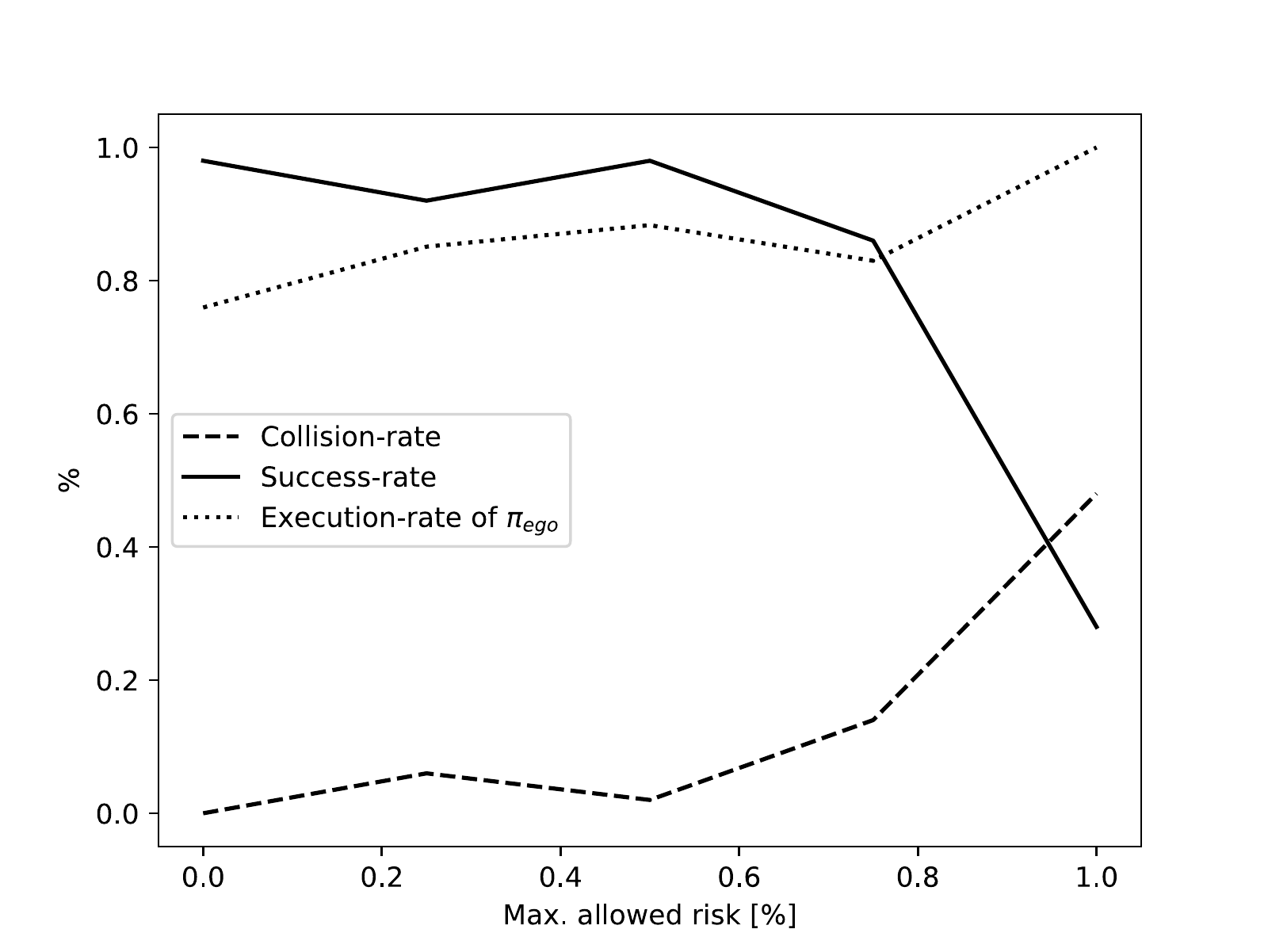
	\caption{The collison- and success-rate of the actual world plotted over the maximum collision-rate threshold $\rho_{max}$.
			 If the collision-rate averaged over all counterfactual worlds is below the maximum collision-rate $\rho_{max}$, the learned policy is executed in the actual world.
			 Otherwise, the ego vehicle is controlled by a lane-following model.}
	\label{fig:rho_succ_over_max_risk}
	\vspace{-0.15cm}
\end{figure}

Figure \ref{fig:influence_heatmap} shows the influence heatmap for the scenario shown in Figure \ref{fig:virtual_worlds} using the mean absolute displacement (MAD) comparing the vehicle's states of the counterfactual worlds to the actual world.
The x-axis represents the counterfactual worlds $W^{(\cdot)}_\mathcal{P}$ in which policies have been exchanged and the y-axis shows the deviation using the MAD of the vehicle's states to the actual world $W$.
For example, the second column in Figure \ref{fig:influence_heatmap} represents the counterfactual world $W^{(1)}_\mathcal{P}$ and the second row shows the MAD for the vehicle $v_1$ over the counterfactual worlds $W^{(j)}_\mathcal{P}$.
As can be seen, replacing the policy of vehicle $v_1$ has a large influence on vehicle $v_0$.
This is expected, as vehicle $v_0$ is controlled by the IDM model and $v_1$ is the leading vehicle.
Vehicle $v_2$ is the leading vehicle of $v_1$ and, thus, has a large impact on vehicle $v_1$.
Vehicle $v_3$ is controlled by the learned policy $\pi_\chi(a | s)$.
As can be seen, all other vehicles have an influence on the learned policy of vehicle $v_3$.
This is an important quantity to observe as this implies that the ego vehicle at least takes all its surrounding vehicles into account.
Vehicle $v_4$ is controlled by the IDM model and, thus, is not being influenced by any other vehicles as there is not leading vehicle in the lane.

Figure \ref{fig:rho_succ_over_max_risk} shows the collision-, execution-, and success-rate of the ego vehicle's policy $\pi_{ego}$ in the actual world $W$ plotted over the maximum allowed collision-rate $\rho_{max}$ of the counterfactual worlds.
If the average conditional collision-rate $P_C$ is lower than the defined collision-threshold $\rho_{max}$, the learned policy $\pi_\chi(a | s)$ of the ego vehicle is executed.
Otherwise, the ego vehicle is controlled by the Mobil model.
We evaluate each of the maximum allowed collision-rates using $250$ episodes.
The success-rate is initially high as the policy is only executed if $\rho_{max}$ is small and, thus, initially no collisions are caused when executing the learned policy.
With a higher maximum allowed collision rate $\rho_{max}$, the execution-rate of the learned policy goes up and the number of collisions increases in the actual world.
This also leads to a significant drop in the success-rate as the goal is not reached as often.
As can be seen, the best results are achieved using a maximum allowed collision-rate $\rho_{max}=0$.
At $\rho_{max}=0$, the ego vehicle's policy $\pi_{ego}$ is executed $76\%$ of the time, the goal is reached with $99.8\%$, and the policy has a collision-rate of $0\%$.

Using the counterfactual policy evaluation prior to the execution of the policy in the actual world decreases the collision-rate significantly.
Even though the policy is not as often executed, the goal is reached almost every single time as can be seen in Figure \ref{fig:rho_succ_over_max_risk}.
Especially for black-box approaches, such as DNNs, the CPE algorithm provides insights on the performance, generalization, and safety prior to execution.
Thus, increasing the applicability of these approaches in safety-critical applications.
It can be seen whether a policy generalizes well judging on how well it can handle the counterfactual worlds.

%% file: figures/eval_virtual_worlds.pdf_tex
\begingroup%
  \makeatletter%
  \providecommand\color[2][]{%
    \errmessage{(Inkscape) Color is used for the text in Inkscape, but the package 'color.sty' is not loaded}%
    \renewcommand\color[2][]{}%
  }%
  \providecommand\transparent[1]{%
    \errmessage{(Inkscape) Transparency is used (non-zero) for the text in Inkscape, but the package 'transparent.sty' is not loaded}%
    \renewcommand\transparent[1]{}%
  }%
  \providecommand\rotatebox[2]{#2}%
  \newcommand*\fsize{\dimexpr\f@size pt\relax}%
  \newcommand*\lineheight[1]{\fontsize{\fsize}{#1\fsize}\selectfont}%
  \ifx\svgwidth\undefined%
    \setlength{\unitlength}{460.79998779bp}%
    \ifx\svgscale\undefined%
      \relax%
    \else%
      \setlength{\unitlength}{\unitlength * \real{\svgscale}}%
    \fi%
  \else%
    \setlength{\unitlength}{\svgwidth}%
  \fi%
  \global\let\svgwidth\undefined%
  \global\let\svgscale\undefined%
  \makeatother%
  \begin{picture}(1,0.75000003)%
    \lineheight{1}%
    \setlength\tabcolsep{0pt}%
    \put(0,0){\includegraphics[width=\unitlength,page=1]{eval_virtual_worlds.pdf}}%
  \end{picture}%
\endgroup%

%% file: figures/eval_influence_heatmap.pdf_tex
\begingroup%
  \makeatletter%
  \providecommand\color[2][]{%
    \errmessage{(Inkscape) Color is used for the text in Inkscape, but the package 'color.sty' is not loaded}%
    \renewcommand\color[2][]{}%
  }%
  \providecommand\transparent[1]{%
    \errmessage{(Inkscape) Transparency is used (non-zero) for the text in Inkscape, but the package 'transparent.sty' is not loaded}%
    \renewcommand\transparent[1]{}%
  }%
  \providecommand\rotatebox[2]{#2}%
  \newcommand*\fsize{\dimexpr\f@size pt\relax}%
  \newcommand*\lineheight[1]{\fontsize{\fsize}{#1\fsize}\selectfont}%
  \ifx\svgwidth\undefined%
    \setlength{\unitlength}{460.79998779bp}%
    \ifx\svgscale\undefined%
      \relax%
    \else%
      \setlength{\unitlength}{\unitlength * \real{\svgscale}}%
    \fi%
  \else%
    \setlength{\unitlength}{\svgwidth}%
  \fi%
  \global\let\svgwidth\undefined%
  \global\let\svgscale\undefined%
  \makeatother%
  \begin{picture}(1,0.75000003)%
    \lineheight{1}%
    \setlength\tabcolsep{0pt}%
    \put(0,0){\includegraphics[width=\unitlength,page=1]{eval_influence_heatmap.pdf}}%
  \end{picture}%
\endgroup%

%% file: figures/eval_plot.pdf_tex
\begingroup%
  \makeatletter%
  \providecommand\color[2][]{%
    \errmessage{(Inkscape) Color is used for the text in Inkscape, but the package 'color.sty' is not loaded}%
    \renewcommand\color[2][]{}%
  }%
  \providecommand\transparent[1]{%
    \errmessage{(Inkscape) Transparency is used (non-zero) for the text in Inkscape, but the package 'transparent.sty' is not loaded}%
    \renewcommand\transparent[1]{}%
  }%
  \providecommand\rotatebox[2]{#2}%
  \newcommand*\fsize{\dimexpr\f@size pt\relax}%
  \newcommand*\lineheight[1]{\fontsize{\fsize}{#1\fsize}\selectfont}%
  \ifx\svgwidth\undefined%
    \setlength{\unitlength}{460.79998779bp}%
    \ifx\svgscale\undefined%
      \relax%
    \else%
      \setlength{\unitlength}{\unitlength * \real{\svgscale}}%
    \fi%
  \else%
    \setlength{\unitlength}{\svgwidth}%
  \fi%
  \global\let\svgwidth\undefined%
  \global\let\svgscale\undefined%
  \makeatother%
  \begin{picture}(1,0.75000003)%
    \lineheight{1}%
    \setlength\tabcolsep{0pt}%
    \put(0,0){\includegraphics[width=\unitlength,page=1]{eval_plot.pdf}}%
  \end{picture}%
\endgroup%

%% file: conclusion.tex
\section{Conclusion}
\label{sec:Conclusion}
This work introduces a counterfactual policy evaluation (CPE) algorithm that evaluates a policy prior to its execution in the real world.
To perform such an evaluation, counterfactual worlds are introduced that evolve differently from the actual world due to exchanged behavior policies of the other vehicles.
Using the evolutions of these worlds, we can reason on how well a policy performs and if it generalizes well enough to handle all counterfactual worlds.
A maximum allowed collision-rate threshold is used that defines if a policy should be executed in the actual world.
If the collision-rate is above the threshold, the policy is not executed in the actual world and the ego vehicle uses a lane-following model instead.
Besides that, the CPE also makes causal relations between the vehicle visible and the influence of the vehicles of each other become evident.
Especially, when using deep neural network it is important to see what and which vehicles the policy takes into account.
Overall, the proposed method increases the applicability and understanding of policies in decision-making for autonomous vehicles significantly.
In future work, the policies of multiple vehicles could be replaced to obtain even deeper insights on the causal relations between the vehicles.
And the field of counterfactual explanations -- also in the view of assurance cases -- could be investigated further.